\documentclass[sigconf]{acmart}

\AtBeginDocument{%
  \providecommand\BibTeX{{%
    \normalfont B\kern-0.5em{\scshape i\kern-0.25em b}\kern-0.8em\TeX}}}

\setcopyright{acmcopyright}
\copyrightyear{2025}
\acmYear{2025}
\acmDOI{XXXXXXX.XXXXXXX}

\acmConference[MM '25]{33rd ACM International Conference on Multimedia}{October 27-October 31, 2025}{Dublin, Ireland}
\acmPrice{15.00}
\acmISBN{978-1-4503-XXXX-X/25/10}

\usepackage{graphicx}
\usepackage{multirow}
\usepackage{enumitem}
\usepackage{bbding}
\usepackage[capitalize]{cleveref}
\crefname{section}{Sec.}{Secs.}
\Crefname{section}{Section}{Sections}
\Crefname{table}{Table}{Tables}
\crefname{table}{Tab.}{Tabs.}

\usepackage{mathrsfs}
\usepackage{booktabs}
\usepackage{multirow}
\usepackage{multicol}
\usepackage{color}
\usepackage{hyperref}
\usepackage{float}
\usepackage{subfig}
\usepackage{colortbl}
\usepackage{rotating}
\usepackage{ulem}

\definecolor{LightCyan}{rgb}{0.88,1,1}
\definecolor{iccvblue}{rgb}{0.21,0.49,0.74}
\definecolor{sgreen}{RGB}{30, 150, 30}

\begin{document}

\title{HOLA: Enhancing Audio-visual Deepfake Detection via Hierarchical Contextual Aggregations and Efficient Pre-training}

\author{Xuecheng Wu$\dagger$}
\email{wuxc3@stu.xjtu.edu.cn}
\affiliation{%
  \institution{Xi'an Jiaotong University}
  \city{Xi'an}
  \country{China}
}

\author{Danlei Huang$\dagger$}
\email{forsummer@stu.xjtu.edu.cn}
\affiliation{%
  \institution{Xi'an Jiaotong University}
  \city{Xi'an}
  \country{China}
}

\author{Heli Sun*$\ddagger$}
\email{hlsun@xjtu.edu.cn}
\affiliation{%
  \institution{Xi'an Jiaotong University}
  \city{Xi'an}
  \country{China}
}

\author{Xinyi Yin$\dagger$}
\email{yinxinyi@stu.zzu.edu.cn}
\affiliation{%
  \institution{Zhengzhou University}
  \city{Zhengzhou}
  \country{China}
}

\author{Yifan Wang}
\email{wangyfan@mail.ustc.edu.cn}
\affiliation{%
  \institution{University of Science and Technology of China}
  \city{Hefei}
  \country{China}
}

\author{Hao Wang}
\email{haowangx@xjtu.edu.cn}
\affiliation{%
  \institution{Xi'an Jiaotong University}
  \city{Xi'an}
  \country{China}
}

\author{Jia Zhang}
\email{zhang\_jia@stu.xjtu.edu.cn}
\affiliation{%
  \institution{Xi'an Jiaotong University}
  \city{Xi'an}
  \country{China}
}

\author{Fei Wang}
\email{fei_wang@mail.dlut.edu.cn}
\affiliation{%
  \institution{Dalian University of Technology}
  \city{Dalian}
  \country{China}
}

\author{Peihao Guo}
\email{3121132005@stu.xjtu.edu.cn}
\affiliation{%
  \institution{Xi'an Jiaotong University}
  \city{Xi'an}
  \country{China}
}

\author{Suyu Xing}
\email{xingsuyu@stu.xjtu.edu.cn}
\affiliation{%
  \institution{Xi'an Jiaotong University}
  \city{Xi'an}
  \country{China}
}

\author{Junxiao Xue$\ddagger$}
\email{xuejx@zhejianglab.cn}
\affiliation{%
  \institution{Zhejiang Lab}
  \city{Hangzhou}
  \country{China}
}

\author{Liang He}
\email{lhe@xjtu.edu.cn}
\affiliation{%
  \institution{Xi'an Jiaotong University}
  \city{Xi'an}
  \country{China}
}

\renewcommand{\shortauthors}{Wu et al.}

\begin{abstract}
Advances in Generative AI have made video-level deepfake detection increasingly challenging, exposing the limitations of current detection techniques. In this paper, we present HOLA, our solution to the Video-Level Deepfake Detection track of 2025 1M-Deepfakes Detection Challenge. Inspired by the success of large-scale pre-training in the general domain, we first scale audio-visual self-supervised pre-training in the multimodal video-level deepfake detection, which leverages our self-built dataset of 1.81M samples, thereby leading to a unified two-stage framework. To be specific, HOLA features an iterative-aware cross-modal learning module for selective audio-visual interactions, hierarchical contextual modeling with gated aggregations under the local-global perspective, and a pyramid-like refiner for scale-aware cross-grained semantic enhancements. Moreover, we propose the pseudo supervised singal injection strategy to further boost model performance. Extensive experiments across expert models and MLLMs impressivly demonstrate the effectiveness of our proposed HOLA. We also conduct a series of ablation studies to explore the crucial design factors of our introduced components. Remarkably, our HOLA ranks 1st, outperforming the second by 0.0476 AUC on the TestA set.
\end{abstract}



\maketitle

\renewcommand{\thefootnote}{\fnsymbol{footnote}}
\footnotetext[1]{Corresponding author.}
\footnotetext[2]{Equal contributions.}
\footnotetext[3]{Equal advice.}

\section{Introduction}
\label{sec:intro}

Generative AI technologies present new opportunities for creating immersive content in various multimedia fields, including VR, cinematic production, and communications. However, their misuse can contribute to the rise of deepfake-related issues, leading to serious social security risks such as online fraud, defamation, and fake news. As a result, developing effective deepfake detection systems is crucial to ensure the reliability of multimedia applications.

In light of this, the 2025 1M-Deepfakes Detection Challenge has brought promising advancements in building powerful and robust deepfake detection systems. The new AV-Deepfake1M++ dataset~\cite{11_avdeepfake1M++}, built upon the AV-Deepfake1M dataset~\cite{26_avdeepfake1m}, contains over 2M samples and thousands of speakers. Regarding its construction procedure, AV-Deepfake1M++ is generated by modifying audio information through word-level \textit{deletion}, \textit{insertion}, and \textit{replacement} cross-grained manipulations, and then aligning fine-grained elements such as lip movements and facial actions with the generated audio and corresponding visual content \cite{26_avdeepfake1m,cai2022you}. The impressive expansion in both quality and quantity presents new challenges for current SOTA deepfake detectors, particularly in capturing audio-visual correlations in the deepfake contexts. In this work, we focus on Track 1, \textit{i.e.}, Video-Level Deepfake Detection, where the task goal is to determine whether a given audio-visual sample, containing a single speaker, is a deepfake or a real video.

Given the above challenges, expert visual-only deepfake detectors~\cite{17_wang2022ffr_fd,ganguly2022visual,murphy2023face} or visual-language models (VLMs)~\cite{Llava-onevision,ViC-Bench,TokenFocus-VQA} often fail to adequately capture the complex deepfake cues embedded in the audio modality. Therefore, we naturally decide to approach the issues from the perspective of audio-visual joint modeling, primarily focusing our design efforts on the aggregations of cross-modal correlations. Most of the existing cross-modal deepfake detection methods~\cite{21_wang2024audio,22_cozzolino2023audio,23_zhang2024ava_cl} typically rely on a limited-scale dataset for specialized design, which often results in models that cannot learn sufficiently generalizable deepfake-related representations, thereby failing to fully unleash the potential of advanced architectures. In the general vision and language domains, large-scale pre-training with scaling properties, followed by fine-tuning on downstream tasks such as multimodal emotion analysis~\cite{wucvpr,wu2025towards,liangyuPTSR} and action recognition~\cite{rcm}, has successfully leveraged the full potential of advanced model designs, leading to significant performance.

\begin{figure*}[t!]
\centering
\includegraphics[width=\linewidth]{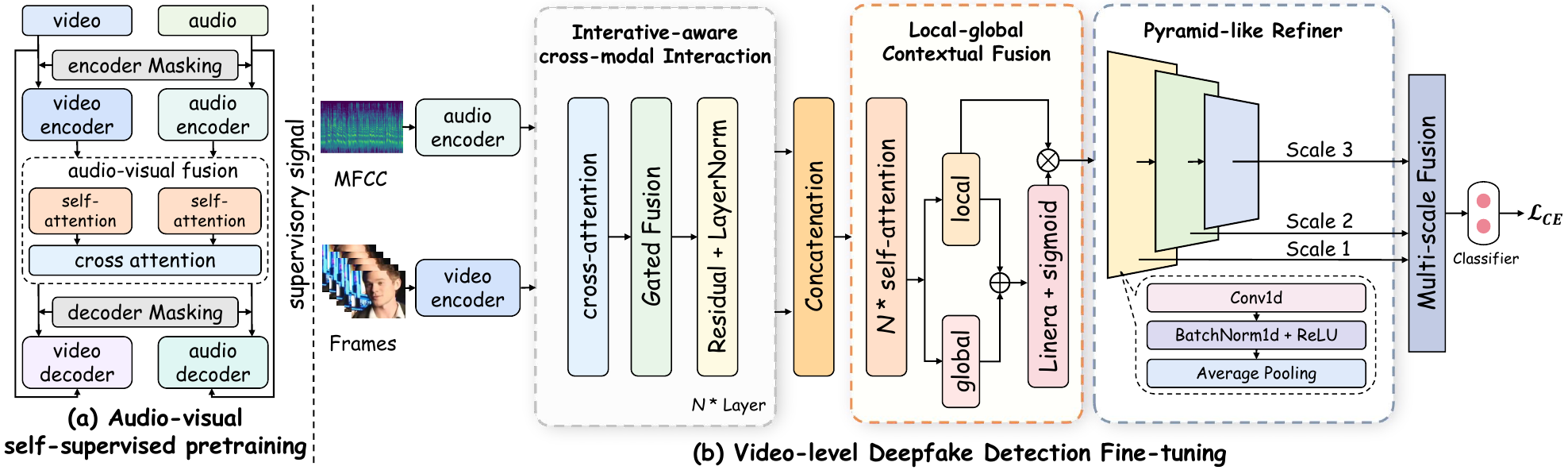}
\vspace{-2.0em}
\caption{The overall illustrations of introduced HOLA , in which our core design is three hierarchical fusion components.}
\vspace{-1.0em}
\label{fig:pipelines}
\end{figure*}

Inspired by such tremendous success, this work proposes a two-stage joint learning framework termed HOLA (Fig.~\ref{fig:pipelines}) for video-level deepfake detection. Specifically, during pre-training, we follow the design of~\cite{hicmae,wucvpr} to construct transformer-based encoders for both audio-visual modalities. We then plainly design an audio-visual fusion module consisting of the attention-based components to capture the correlated information. Additionally, we introduce the audio-visual dual masking strategy to balance the performance and training costs with an acceptable drops. Building upon this, we construct a massive dataset consisting of 1.81M samples sourced from the speaker recognition datasets~\cite{fan2020cnceleb,hdtf} to support scalable pre-training, thereby establishing a solid foundation for downstream deepfake detection. During deepfake fine-tuning, leveraging the audio-visual encoders which learn generalized facial representations, we design three distinct components from the hierarchical contextual perspective to fully aggregate cross-modal correlations for capturing deepfake cues. First, we propose an iterative-aware cross-modal learning module to perform selective interactions, thereby suppressing intra-modal redundancies and enhancing complementary forgery cues. Based on the enhanced features, we then develop a local-global contextual fusion module with gated aggregations to achieve a comprehensive capture of both local tampering artifacts and global semantic inconsistencies. Finally, we introduce a pyramid-like refiner to perform scale-aware cross-grained semantic enhancements, which is coupled with a classifier to output the final predictions. Moreover, we introduce a pseudo-supervised signal injection strategy, which boosts the fine-tuning scale by extracting the absolutely confident samples from each iteration, leading to further performance gains. Extensive experimental results across AV-Deepfake1M++~\cite{11_avdeepfake1M++} subsets demonstrate the effectiveness of our HOLA. Moreover, we evaluate the performance of advanced VLMs in this scenario to further validate the superiority of our expert model. We also perform ablation studies to explore the critical design factors of our proposed components. Notably, we achieve the first place in this video-level deepfake detection track, outperforming the second place by 0.0476 AUC on the TestA set.

In addition to advancing deepfake detection, our HOLA can also inspire advancements in audio-visual scene understanding and omni-modal language models~\cite{luo2025openomni,guo2025streamuni}. In summary, our main contributions are three-fold: \textbf{(i)}.~We efficiently perform data scaling in learning general facial representations for audio-visual deepfake detection, constructing 1.81M pre-training samples, thereby establishing a two-stage training paradigm for impressive performance. \textbf{(ii)}.~We propose the iterative-aware cross-modal learning module,  local-global contextual fusion module, and pyramid-like refiner to perform hierarchical contextual aggregations, facilitating the capture of deepfake-related cues. \textbf{(iii)}.~We introduce the pseudo-supervised signal injection strategy to globally enhance performance. Extensive experiments across AV-Deepfake1M++ subsets demonstrate the effectiveness of our HOLA. A series of ablation studies are also conducted to explore the key factors of our design.

\section{Related Work}
\label{sec:Related Work}

\subsection{Deepfake Detection Datasets}
Existing deepfake detection datasets can be generally classified into unimodal and multimodal ones. Unimodal datasets are predominantly composed of visual modality, for instance, FaceForensics++ \cite{1_faceforensics++} provides over 1.8M images extracted from 1,000 real and manipulated videos. Celeb-DF \cite{2_celeb} comprises 590 real and 5,639 forged videos of celebrities. Deepfake-Synthetic-20K \cite{4_Deepfake-Synthetic-20K} offers 20,000 high-resolution synthetic face images generated using StyleGAN-2~\cite{25_stylegan2}. Multimodal datasets typically contain two or more modalities among images, videos, and audio, making them suitable for cross-modal deepfake detection research. FakeAVCeleb \cite{7_fakeavceleb} contains 500 real videos and 19,500 fake videos. ForgeryNet \cite{8_forgerynet} includes 2.9M images and 221,247 videos, covering 7 image-level and 8 video-level forgery techniques. DF40 \cite{9_df40} encompasses 40 deepfake generation techniques, featuring over 0.1M forged video clips and more than 1M fake images. DDL \cite{10_ddl} offers over 1.8M fake samples across image, video, and audio modalities. The dataset used in this work, AV-Deepfake1M++ \cite{11_avdeepfake1M++}, is the currently largeset audio-visual multimodal deepfake detection dataset comprising over 2M high-quality video clips and covering over 2,000 speakers.

\subsection{Deepfake Detection Methods}
Existing approaches in deepfake detection can be categorized into visual-based and multimodal ones. Specifically, visual-based methods can be further divided into image-level and video-level detection. Image-level detection relies on digital forensics and the identification of visual artifacts. Lugstein et al. \cite{12_lugstein2021prnu} introduce PRNU from the forensics domain into deepfake detection. Detectify \cite{14_detectify} combines ELA with CNNs for robust detection. Zhao et al. \cite{15_Zhao_2021_CVPR} regard this detection task as fine-grained classification and design a multi-attention model. Nguyen et al. \cite{16_Nguyen_2024_CVPR} propose a localized artifact attention network under the multi-task manner. Video-level detection further exploits temporal cues to identify inconsistencies over time. Zheng et al. \cite{18_Zheng_2021_ICCV} propose an end-to-end framework that leverages temporal coherence. Ru et al. \cite{19_ru2021bita} introduce Bita-Net, a dual-path architecture covering temporal consistency and semantic details. In multimodal video-level deepfake detection, AVFF \cite{20_oorloff2024avff} proposes a method to capture synchronization between audio and visual streams. ART-AVDF \cite{21_wang2024audio} designs phoneme encoders and lip encoders to capture audio-visual consistency. Zhang et al. \cite{23_zhang2024ava_cl} propose AVA-CL, which leverages intrinsic correlations between audio and video cues. Cozzolino et al. \cite{22_cozzolino2023audio} extract speaker-related multimodal features and develop a point-of-interest-based detector.

\section{Methodology}
\label{sec:Mthods}

\subsection{Audio-visual Self-supervised Pre-training}
\label{3.2}

\subsubsection{Model Architecture}
 
We introduce a self-supervised pretraining framework for learning expressive video and audio encoders through joint audio-visual modeling. Inspired by~\cite{hicmae}, the framework consists of four main components: a visual encoder, an audio encoder, a cross-modal fusion encoder, and modality-specific decoders. Raw video and audio inputs are first transformed into discrete token sequences via cube embedding and spectrogram embedding modules, respectively. Positional encodings are then added, resulting in the formal inputs $ \overline{X}_v \in \mathbb{R}^{N_v \times C} $ and $ \overline{X}_a \in \mathbb{R}^{N_a \times C} $.

To improve training efficiency and robustness, high masking ratios are applied. Specifically, the video branch uses tube masking with a 90\% masking ratio, while the audio branch adopts 80\% random masking. After masking, the remaining visible tokens are processed by the transformers-based encoders to extract latent representations, which are then forwarded to a fusion module for extracting cross-modal features.
For effective fusion, we introduce a interaction module composed of self-attention and cross-attention components. In this fusion module,multi-head self-attention (MHSA) first captures intra-modal dependencies:
\begin{align}
\text{MHSA}(F_{m_c}) &= \text{Concat}(h_1, ..., h_H) W^O, \\
 h_i &= \text{Softmax} \left( \frac{Q_i K_i^\top}{\sqrt{d_k}} \right) V_i,
\end{align}
where $ F_{m_c} $ denotes the input features of the current modality, $ h_i $ is the attention output of the $ i $-th head, $ H $ is the number of heads, and $ W^O $ is the output projection matrix. $ Q_i, K_i, V_i $ are the \textit{query}, \textit{key}, and \textit{value} matrices for the $ i $-th head, respectively, and $ d_k $ is the dimensionality of the \textit{key} features. Afterwards, multi-head cross-attention (MHCA) is utilized to enable inter-modal interactions by treating visual tokens as \textit{query} and audio tokens as \textit{key} and \textit{value}, and vice versa. For modality $ m_c,m_o \in \{v, a\} $ :
\begin{align}
\text{MHCA}(F_{m_c}, F_{m_o}) &= \text{Concat}(h_1, ..., h_H) W^O, \\
h_i &= \text{Softmax} \left( \frac{Q_i^{m_c} (K_i^{m_o})^\top}{\sqrt{d_k}} \right) V_i^{m_o},
\end{align}
where \( F_{m_c}, F_{m_o} \in \{F_v, F_a\} \) denote the current modality features and the other one, respectively. And $ Q_i^{m_c}, K_i^{m_o}, V_i^{m_o} $ are the $m_c$ query and $m_o$ key and value matrices.

Following the fusion stage, we concatenate the encoded features with learnable masked tokens to construct the inputs for each modality-specific decoder, which reconstructs the original signal from the partial observation. To mitigate massive computational overhead, we introduce the audio-visual dual-masking strategy from AVF-MAE++~\cite{wucvpr}, which impressively reduces the token number of both decoders. Given encoder mask $ M^{en} $ and decoder mask $ M^{de} $, the decoder input is defined as:
\begin{equation}
F_{m_c}^t = F_{m_o \to m_c} \cup \{ M_i^{m_c} \mid i \in M_{m_c}^{de} \},
\end{equation}
where $ F_{{m_o} \to m_c} $ represents visible features transferred from the opposite modality, while $ M_i^{m_c} $ are learnable masked tokens at positions indexed by $ M_{m_c}^{de} $. To be specific, the video decoder employs running cell masking \cite{rcm} to periodically retain spatial blocks along the temporal dimension, enhancing the model's capacity to capture long-range dependencies. Meanwhile, the audio decoder retains its random masking strategy to exploit short-range contextual cues. Both decoders operate with a masking ratio of 50\%.

Regarding model optimization, the primary objective is to minimize the reconstruction error at the masked locations, utilizing a modality-specific Mean Squared Error (MSE) loss function, which can be formulated as:
\begin{equation}
    \mathcal{L}_{\text{MSE}}^{m_c} = \frac{1}{(1 - \lambda_{m_c}) \cdot N_{m_c}} \sum_{i \in M_{m_c}^{de} \cap M_{m_c}^{en}} \left\| X_i^{m_c} - \hat{X}_i^{m_c} \right\|^2,
\end{equation}
where $ X_i^{m_c} $ and $ \hat{X}_i^{m_c} $ are the ground truth and reconstructed features at position $ i $, $ M_{m_c}^{de} \cap M_{m_c}^{en} $ indicates the masked positions used for loss computation, $ N_{m_c} $ is the number of tokens, and $ \lambda_{m_c} $ is the masking ratio for modality $ m_c $. As a result, the total loss is computed as the sum of the reconstruction losses from both audio and video modalities:
\begin{equation}
    \mathcal{L}_{\text{total}} = \mathcal{L}_{\text{MSE}}^a + \mathcal{L}_{\text{MSE}}^v.
\end{equation}

\subsubsection{Data}

To better support the pretraining of audio and visual encoders, we construct a large-scale hybrid pretraining dataset by integrating diverse sources, including CelebV-HQ \cite{zhu2022celebv}, the CNC-AV series \cite{fan2020cnceleb}, HDTF \cite{hdtf}, and MSD-Wild-DB \cite{msdwild}, further expanding upon the dataset foundation established in AVF-MAE++~\cite{wucvpr}, which enriches the variety of audio-visual content and ensures broader coverage for model training. A unified preprocessing pipeline~\cite{fan2020cnceleb} is employed to filter and segment raw videos, removing redundant content and standardizing video formats and durations. This process yields a large-scale self-supervised dataset comprising 1.81M videos.

We adopt the same pretraining settings as the base version of AVF-MAE++ \cite{wucvpr} to conduct large-scale audio-visual self-supervised learning. As a result, we obtain two modality-specific transformers-based encoders that can serve as visual and audio backbones with strong representational capacity. Trained on such diverse and large-scale data, these audio-visual encoders are subsequently integrated into our downstream fine-tuning for the video-level audio-visual deepfake detection.

\subsection{Downstream Deepfake Detection Tuning}
\label{3.3}

\subsubsection{Pre-processing}
For the video modality, we uniformly extract $ T $ frames and resize each frame to a consistent size of $ H \times W $, leading to video features $ F_v \in \mathbb{R}^{T \times H \times W \times C} $ , where $C$ is the feature dimension. Regarding the audio modality, we process them by extracting Mel-frequency cepstral coefficients (MFCC). The audio signal is divided into multiple time segments, and MFCC coefficients are calculated for each segment, resulting in audio features $ F_a \in \mathbb{R}^{T \times F} $, where $ F $ is the number of MFCC coefficients for each time segment. The audio time steps $ T $ are aligned with the number of video frames $ T $ through transformations to ensure temporal consistency between audio and video branches.

\subsubsection{Backbone Feature Extractions}

To convert the processed samples into a format suitable for processing by the transformers-based pretrained encoders, we follow \cite{wucvpr} and \cite{hicmae}, using an embedding-based approach to convert the features of both modalities into token sequences, resulting in video features $ F_v \in \mathbb{R}^{L_v \times C} $ and audio features $ F_a \in \mathbb{R}^{L_a \times C} $, where $ L_v $ and $ L_a $ denote the lengths of the video and audio token sequences, respectively.

After embedding, the token sequences are fed into their respective pretrained encoders. Following the transformers-based encoding, we employ a one-dimensional convolution block followed by a linear projection, mapping both video and audio features to a unified sequence length $L$, leading to the final features \( F_v, F_a \in \mathbb{R}^{L \times C} \).

\subsubsection{Iterative-aware Cross-modal Interaction}
To facilitate deep interactions between video and audio modalities, we design an Iterative-aware Cross-modal Interaction Module, which enhances video and audio features by stacking three times of interaction modules. To be specific, in each interaction module, we first employ bidirectional cross-attention to capture cross-modal relationships between features, in which the current features $ F_{m_c} $ serve as queries, while the other features $ F_{m_o} $ act as contexts to update $ F_{m_c} $. After the cross-attention interactions, a gating mechanism is applied, consisting of two linear layers with a ReLU activation in between to generate fusion weights. This gating mechanism controls the fusion ratio of the features, effectively avoiding a blind summation. The overall process can be formulated as:
\begin{equation}
G = \sigma(W_g \cdot [F_{m_c}, \text{MHCA}(F_{m_c}, F_{m_o})]),
\end{equation}
where $ W_g $ is the learnable weight matrix, and $ \sigma $ represents the sigmoid function. The residual connections are then added to the fused features $ F_u $, which can be represented as:
\begin{equation}
F_u = (1 - G) \cdot F_{m_c} + G \cdot \text{MHCA}(F_{m_c}, F_{m_o}).
\end{equation}

In this way,  the final updated output after residual connection and layer normalization can be calculated as:
\begin{equation}
\hat{F_{m_c}} = \text{LayerNorm}(F_{m_c} + F_{\text{u}}).
\end{equation}

The three stacked layers progressively deepen the interaction and fusion of the audio-visual features. In each layer, the video and audio features are first updated, and long-range dependencies are better captured through attention-based components, leading to interacted outputs \( \hat{F_v}, \hat{F_a} \in \mathbb{R}^{L \times C} \).

\subsubsection{Local-Global Contextual Fusion}
To effectively supplement semantic features with local and global aware abilities, we design the Local-Global Contextual Fusion Module, based on Hierarchical context-awareness and gated-weighting design principles. This module extracts global and local information from the context of the audio-visual modalities and achieves high-quality joint fusion through a gating mechanism.

Specifically, after iteratively interacting, we concatenate the two resulting tensors to obtain \( F_c \in \mathbb{R}^{2L \times C} \). To enhance the global semantic information, we introduce a global representation \textbf{[CLS]} token to capture the cross-modal global-aware context, and insert it at the beginning of $ F_c $, leading to a new feature \( F_c' \in \mathbb{R}^{(2L + 1) \times C} \). Next, $ F_c' $ is passed through $ N $ layers of self-attention to perform comprehensive contextual modeling. The first element of the output feature corresponds to the global representation, \textit{i.e.}, \textbf{[CLS]} token.

To better leverage the local and global features, we explicitly separate them and utilize the gating mechanism again. This gating consists of linear and activation layers,  controlling the weighted ratio between the local features $ F_l $ and global features $ F_g $, which is then deployed to generate the weighted fused feature \( F_f \in \mathbb{R}^{2L \times C} \): 
\begin{equation}
F_f = F_l \cdot \sigma(\hat{W_g} \cdot Concat(F_l, F_g)).
\end{equation}

\subsubsection{Pyramid-like Refiner}

To enhance multi-scale perceptual ability and capture both short-term dynamic details and long-range contextual dependencies, while simultaneously adjusting feature dimensions to shape the final outputs, we design the Pyramid-like Refiner, which can extract fine-grained semantics across multiple temporal scales and fuse them to generate a discriminative global feature for binary classification tasks.

We introduce three scales, each including a convolutional block with Conv1D, BatchNorm1D, ReLU activation, and average pooling. Each scale uses a stride-2 Conv1D layer to downsample the sequence, progressively reducing its length and capturing features at different scales, thus forming a pyramid structure. Each scale leads to $ F_{s_i} \in \mathbb{R}^{C} $, and the features from three scales is concatenated to generate the global-aware feature $ F_{ms} \in \mathbb{R}^{3C} $, \textit{i.e.},
\begin{align}
F_{ms}  &= \text{Concat}(F_{s_1}, F_{s_2}, F_{s_3}), \\
F_{s_i} &= \text{GAP}(\text{Conv1D}(F_{s_{i-1}})),
\end{align}
where $ F_{s_i} $ is the feature after applying global average pooling (GAP) to the outcome of the 1D convolution block of the previous scale's output. $ F_{s_{i-1}} $ refers to the feature from the previous scale.

After the linear transformations and normalization, the semantic features are aggregated into a unified feature $ F_{s_f} \in \mathbb{R}^{C} $. In the end, a classifier is used to perform binary classification, leading to the final output. Additionally, we employ the standard cross-entropy (CE) loss to minimize the overall model, facilitating the effective fine-tuning of our proposed HOLA on the downstream video-level deepfake detection task.

\subsection{Pseudo Supervised Singal Injection Strategy}
\label{3.4}
To make better use of the unlabeled test subset, we introduce the pseudo-supervised signal injection strategy to further boost model performance. Specifically, after each training iteration, we select all the samples with the confidence score = 1 for both real and fake categories based on the final predictions. These absolute samples are then added to the training set, leading to more samples for next iteration. By updating the training set with pseudo-labels through multiple iterations, HOLA can learn more comprehensively discriminative patterns, enhancing its capability to detect deepfakes.
\section{Experiments}
\label{sec:exprs}

\subsection{Experimental Settings}
\label{4d1}

\subsubsection{Implementation Details}
We follow HiCMAE~\cite{hicmae} and AVF-MAE++~\cite{wucvpr} to perform large-scale pre-training settings. Our downstream model is trained for 30 epochs with batch size = 64 on a machine with 8 $\times$ NVIDIA A100 GPUs (80GB) using PyTorch 2.5.1~\cite{2019pytorch}. We deploy AdamW~\cite{adamw} with weight decay 2$e$-2 and the initial learning rate 1.5$e$-4 using cosine schedule. For the visual branch, we uniformly sample 16 frames. Each frame is then resized to align the short-side length to 224 pixels. Additionally, we use the \href{http://www.ffmpeg.org/}{FFmpeg} toolbox to pre-process visual samples. For the audio branch, we first transform the audio into monophonic and set the sampling rate at 16,000 Hz. The audio is then converted into Numpy arrays and normalized. Next, we deploy the \href{https://librosa.org}{Librosa} library to compute MFCCs, utilizing 32 coefficients. Subsequently, the processed audio is divided into 16 segments as the formal input of our audio branch.

\subsubsection{Datasets}

AV-Deepfake1M++ \cite{11_avdeepfake1M++} is an extended version of AV-Deepfake1M \cite{26_avdeepfake1m} dataset, containing over 2M video samples and covering a larger set of speakers. The dataset is split into four subsets, \textit{i.e.}, Training, Validation, TestA, and TestB.

\subsubsection{Evaluation Metrics}
In the video-level deepfake detection task, AUC is employed as the official metric, with higher values indicating better performance, which can be formulated as:
\begin{equation}
AUC = \frac{\sum_{\textit{ins}_i \in \textit{positive class}} \textit{rank}_{\textit{ins}_i}-\frac{M \times (M+1)}{2}}{M \times N},
\end{equation}
where $M$ and $N$ are the quantity of positive and negative samples. $\textit{rank}_{\textit{ins}_i}$ denotes the rank of a positive sample $\textit{ins}_i$ among all samples, and $\frac{M(M+1)}{2}$ is a correction term for the ideal positive rank sum. To provide more comprehensive evaluations, we also report the mainstream metrics, \textit{i.e.}, Accuracy (ACC), Unweighted Average Recall (UAR), as well as Weighted Average F1 Score (WA-F1).

\subsection{Performance Comparisons}
\label{4d4}

\subsubsection{Expert Models}

\begin{table}[t!]
\centering
\caption{Performace comparisons of advanced baselines and our proposed HOLA in terms of ACC, WA-F1, UAR, and AUC metrics on the video-level deepfake detection track. Note that we highlight the best performance in \textbf{bold} and \underline{underline} the second performance.}
\vspace{-0.5em} 
\setlength{\arrayrulewidth}{0.40pt}
\renewcommand{\arraystretch}{1.25} % 行高
\resizebox{\linewidth}{!}{%
\begin{tabular}{lcccccccc}
\toprule
Method & Venue & ACC~(\%) & WA-F1~(\%) & UAR~(\%) & AUC \\ 
\midrule
ECAPA-TDNN~\cite{new_32_desplanques2020ecapa} & Interspeech & 80.04 & 77.14 & 65.37 & 0.8569\\
Res2Net~\cite{34_gao2019res2net} & TPAMI & 83.10 & 82.78 & 75.88 & 0.9059 \\
CAM++~\cite{38_wang2023cam++}  & Interspeech & 74.52 & 64.28 & 50.81 & 0.5182 \\
ResNet-SE~\cite{hu2018squeeze} & CVPR & 81.88 & 81.35 & 74.25 & 0.8942 \\
PANNS~\cite{27_kong2020panns}  & TASLP & 82.39 & 80.70 & 70.00 & \underline{0.9102} \\

I3D~\cite{43_carreira2017quo}  &  CVPR & 85.00 & 84.76 & \underline{78.97} & 0.9058 \\

C3D~\cite{32_tran2015learning} &  CVPR & 74.31 & 64.03 & 50.63 & 0.6506 \\

Video Swin-T~\cite{19_liu2022video} & CVPR & 76.61 & 68.71 & 54.88 & 0.7560 \\

TimeSformer~\cite{30_bertasius2021space}  & ICML & \underline{86.04} & \underline{85.39} & 77.93 & 0.9097 \\

\rowcolor{iccvblue!20}
\textbf{HOLA} (ours) & \textbf{--} & \textbf{98.63} & \textbf{98.62} & \textbf{98.07} & \textbf{0.9991} \\
\hline
\end{tabular}
}
\vspace{-0.5em}
\label{tab:main-compare}
\end{table}

We compare our HOLA with current state-of-the-art baselines on the \textit{Validation} set of AV-Deepfake1M++~\cite{11_avdeepfake1M++} dataset, as shown in Tab.~\ref{tab:main-compare}. It can be clearly observed that our method achieves the best performance on the video-level deepfake detection track, surpassing the second-best approach by 12.59\% in terms of accuracy (ACC) metric. These experimental results clearly demonstrate that the general representations obtained through large-scale pre-training lead to more stable and robust performance compared to the expert models trained from scratch. Moreover, the significant performance improvement also benefits from our advanced audio-visual correlation learning components and the global-aware pseudo-supervised signal injection strategy, which together help the overall model to capture more discriminative deepfake-related representations.

\subsubsection{Visual-Language Models}

\begin{table}[t!]
\centering
\caption{Performace comparisons of advanced VLMs in terms of ACC, WA-F1, and UAR on video-level deepfake detection.}
\vspace{-0.5em} 
\setlength{\arrayrulewidth}{0.40pt}
\renewcommand{\arraystretch}{1.20}
\resizebox{\linewidth}{!}{%
\begin{tabular}{lcccccccc}
\toprule
Method & Organization & ACC~(\%) & WA-F1~(\%) & UAR~(\%)  \\ 
\midrule

o3-mini-high~\cite{openai2025o3mini} & OpenAI & 46.06 & 48.34 & 50.60 \\ 
Gemini-2.5-pro~\cite{Gemini-2.5-pro-preview-03-25} & Google & 73.30 & 62.01 & 50.00 \\
Gemini-2.5-pro-thinking~\cite{Gemini-2.5-pro-preview-03-25} & Google & \textbf{73.88} & \textbf{62.39} & \textbf{51.00} \\
Claude-3.7-Sonnet-250219 & Anthropic & 29.03 & 18.59 & 49.60 \\
GPT-4o-240513~\cite{openai2024gpt4o} & OpenAI & 60.80 & 59.61 & 47.31\\ 

QVQ-Max-250325~\cite{Qwen-QVQ-72B}  & Alibaba & \underline{73.32} & \underline{62.03} & 50.00  \\ 

Qwen2.5-VL-32B-Instruct~\cite{Qwen2d5-vl}  & Alibaba & 32.40 & 25.79 & 50.55  \\ 
Qwen2.5-VL-72B-Instruct~\cite{Qwen2d5-vl}  & Alibaba & 30.90 & 22.53 & 50.25  \\ 
Grok 4~\cite{grok-4} & xAI & 73.30 & 62.01 & 50.00  \\
Phi-3.5-Vision-Instruct~\cite{48_abdin2024phi} & Microsoft & 50.70 & 53.51 & 48.63  \\
Doubao-Seed-1.6-250615~\cite{Doubao-1.5-vision-pro-32k} & ByteDance & 30.10 & 19.93 & \underline{50.77} \\

\hline
\rowcolor{gray!20}
\textbf{HOLA} (Reference) & \textbf{--} & 98.83 & 98.59 & 97.74 \\
\hline
\end{tabular}
}
\label{tab:main-compare-mllms}
\end{table}

To further quantitatively analyze the performance of advanced large visual-language models (VLMs) on the video-level deepfake detection track, we randomly sample 1,000 instances from the \textit{Validation} set of AV-Deepfake1M++~\cite{11_avdeepfake1M++} dataset and conduct zero-shot inference, as shown in Tab.~\ref{tab:main-compare-mllms}. We additionally include the performance of our proposed HOLA on the same subset for comprehensive comparisons. The extensive experimental results indicate that reasoning-capable MLLMs such as~\cite{Gemini-2.5-pro-preview-03-25,Qwen-QVQ-72B} outperform vanilla MLLMs by a notable margin. For example, Gemini-2.5-Pro-Thinking achieves 42.98\% higher accuracy than Qwen2.5-VL-72B~\cite{Qwen2d5-vl}. However, the overall zero-shot performance of MLLMs still falls short of the expert model, highlighting the necessity of incorporating deepfake-related data samples during pre-training and supervised fine-tuning (SFT) to further enhance their deepfake-related discriminative capabilities.

\subsection{Effectiveness of Pseudo Signal Injection}
\label{4d5}

\begin{figure}[t!]
\centering
\includegraphics[width=\linewidth]{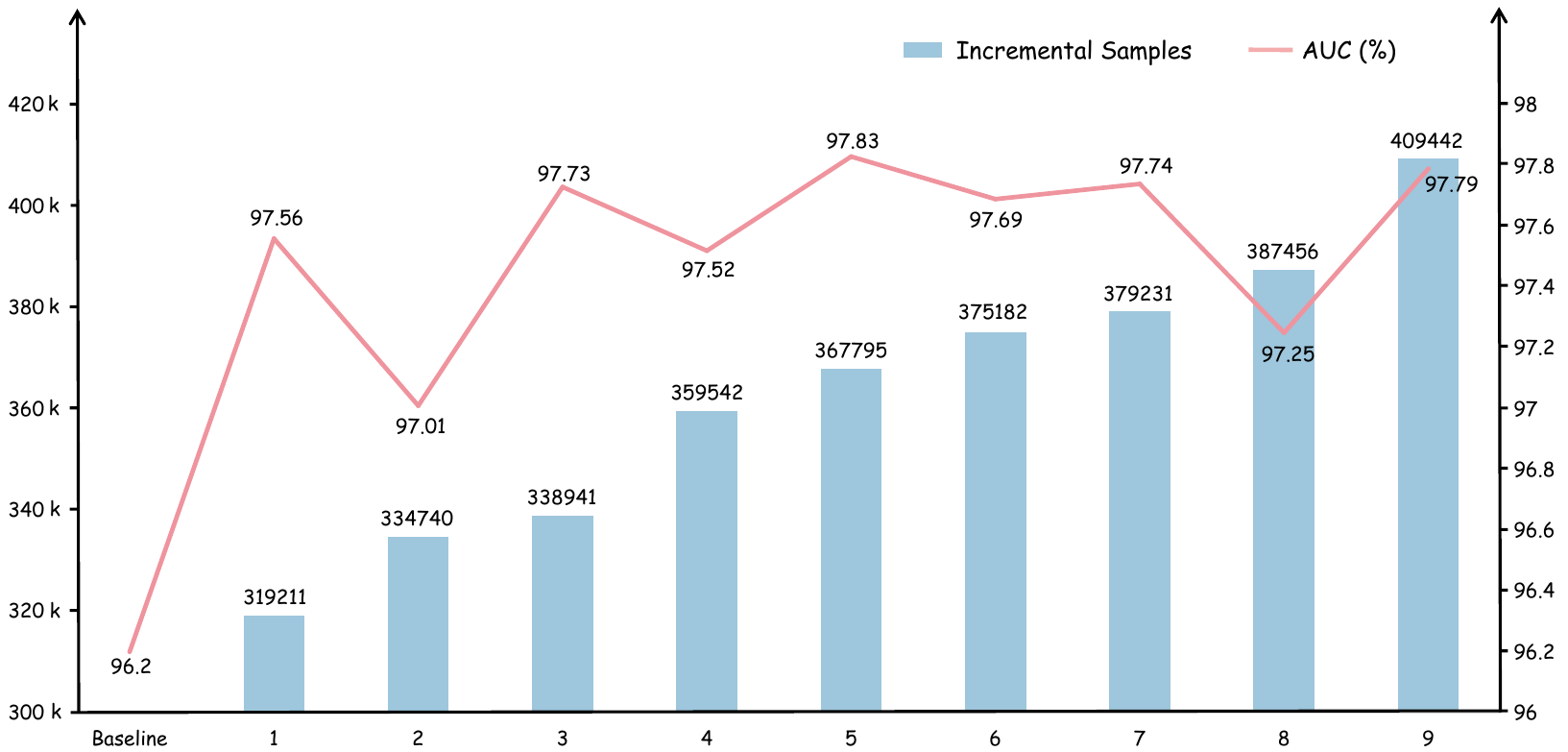}
\vspace{-2.0em}
\caption{The results of pseudo-supervised signal injection.}
\vspace{-1.0em}
\label{pseudo_signal_injection}
\end{figure}

The pseudo-supervised signal injection strategy steadily incorporates absolute confidence samples into deepfake fine-tuning, enlarging the decision boundary of our HOLA to align with low-density regions and facilitating more comprehensive capture of discriminative patterns. As shown in Fig.~\ref{pseudo_signal_injection}, after five iterations, HOLA achieves a peak AUC of 97.83\% on testA, exhibiting 1.69\% AUC improvement over the previous best performance. However, performance begins to decline beyond 5 iterations, likely due to the repeated reinforcement of incorrect pseudo-labels. This can mislead the model’s decision boundary and ultimately degrade performance.

\begin{table}[t!]
\centering
\caption{The ablation studies of introduced components in terms of ACC and WA-F1 on the Validation set.}
\vspace{-0.5em}                   
\setlength{\arrayrulewidth}{0.40pt}
\renewcommand{\arraystretch}{1.1}
\resizebox{\linewidth}{!}{%
\begin{tabular}{lcccc}
\toprule
Method & ACC~(\%)  & WA-F1~(\%)   \\ 
\midrule
Baseline         & 95.27 & 94.91 \\
+ Interative-aware Cross-modal Interaction  & 96.22 & 96.03 \\
+ Local-global Contextual Fusion  & 97.40 & 97.52 \\
+ Pyramid-like Refiner & 98.21 & 98.32 \\
+ Pseudo-supervised Signal Injection Strategy & \textbf{98.63} & \textbf{98.62} \\
\hline
\end{tabular}
}
\vspace{-0.5em}
\label{tab-overall-ablation}
\end{table}

\subsection{Ablation Study}
\label{4d6}

We conduct the overall ablation studies on the \textit{Validation} set, as displayed in Tab.~\ref{tab-overall-ablation}. With the support of our large-scale audio-visual self-supervised pre-training, our baseline model already outperforms other state-of-the-art expert models in the downstream video-level deepfake detection. Building on this, the incorporation of our proposed modules, \textit{i.e.}, iterative-aware cross-modal interaction, local-global contextual fusion, and pyramid-like refiner, further enhances model performance, totally achieving enhancements of 2.94\% ACC and 3.41\% WA-F1. These results validate the effectiveness of our introduced cross-modal correlations learning components. In the end, with the introduction of our pseudo-supervised signal injection strategy, additional deepfake-related learning factors are injected at the global level, leading to more performance gains.

\subsection{Qualitative Analysis}
\label{4d7}
\begin{figure}[t!]
\centering
\includegraphics[width=\linewidth]{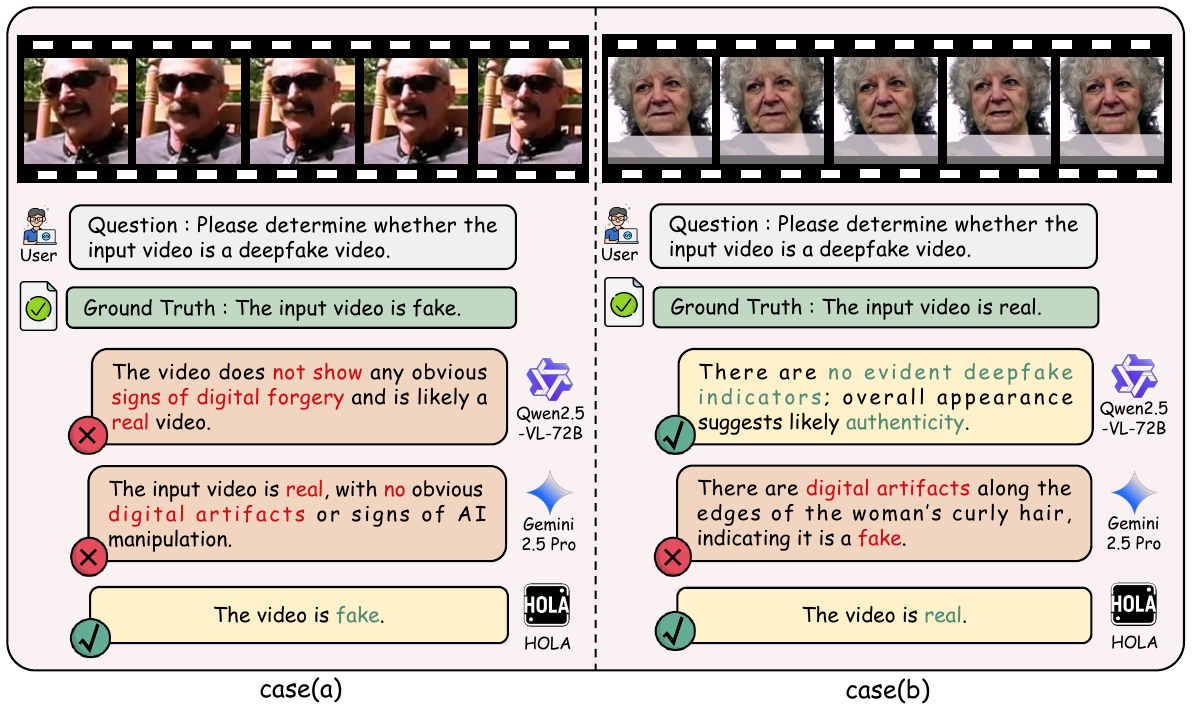}
\vspace{-2.2em}
\caption{Qualitative comparisons of HOLA and SOTA VLMs.}
\vspace{-1.3em}
\label{Qualitative_Analysis}
\end{figure}

To further validate the effectiveness of our HOLA in the real-world scenarios, we compare proposed HOLA and advanced VLMs~\cite{Qwen2d5-vl,Gemini-2.5-pro-preview-03-25} on two representative samples. From Fig.~\ref{Qualitative_Analysis}, we can clearly figure out that despite the strong scene understanding capabilities of VLMs~\cite{Qwen2d5-vl,Gemini-2.5-pro-preview-03-25}, they still often fail to capture the deepfake-related cues without targeted supervised fine-tuning. In contrast, our HOLA can accurately recognize both samples by leveraging the complementary audio-visual cues, exhibiting reliable deepfake detection. These results highlight that unlocking the full generative capability of VLMs requires dedicated architectures and data strategies.

\section{Conclusions}
\label{sec:Conclusions}

In this paper, we introduce a unified two-stage framework termed HOLA for video-level deepfake detection. To be specific, we efficiently scale data by utilizing our self-built pre-training dataset of 1.81M samples to learn general audio-visual facial representations. Building upon this, we propose the iterative-aware cross-modal learning module, local-global contextual fusion module, and pyramid-like refiner to perform hierarchical aggregations, which facilitates the capture of deepfake-related representations. Moreover, we introduce a pseudo-supervised signal injection strategy to further enhance performance at the global level. Extensive qualitative and quantitative experiments across AV-Deepfake1M++ demonstrate the superiority of our HOLA. We also conduct a series of ablation studies to investigate the crucial factors of our design.

\clearpage
\newpage
\bibliographystyle{ACM-Reference-Format}
\bibliography{reference}

\end{document}